# Clarifying System 1 & 2 through the Common Model of Cognition

Brendan Conway-Smith (brendan.conwaysmith@carleton.ca),
Robert L. West (robert.west@carleton.ca)
Department of Cognitive Science, Carleton University
Ottawa, ON K1S5B6 Canada

**Abstract**

There have been increasing challenges to dual-system descriptions of System-1 and System-2, critiquing them as imprecise and fostering misconceptions. We address these issues here by way of Dennett's appeal to use computational thinking as an analytical tool, specifically we employ the Common Model of Cognition. Results show that the characteristics thought to be distinctive of System-1 and System-2 instead form a spectrum of cognitive properties. By grounding System-1 and System-2 in the Common Model we aim to clarify their underlying mechanisms, persisting misconceptions, and implications for metacognition.

## Introduction

This paper re-visits Dennett's (1981) notion that philosophical discussion can benefit from the use of computational modelling. We do this by showing how recent criticisms of the dual-systems view of the mind (System-1 and System-2), can be clarified using the Common Model of Cognition to ground the discussion (Laird, Lebiere & Rosenbloom, 2017).

The terms System-1 and System-2 refer to a dual-system model that ascribes distinct characteristics to what are thought to be opposing aspects of cognition (Wason & Evans, 1974; Stanovich, 1999; Strack & Deutsch, 2004; Kahneman, 2003, 2011). System-1 is considered to be evolutionarily old and characterized as fast, associative, emotional, automatic, and not requiring working memory. System-2 is more evolutionarily recent and thought to be slow, declarative, rational, effortful, and relying on working memory. Kahneman (2003) referred to System-1 as "intuitive" and System-2 as "rational", thus linking them to higher level folk psychology concepts. The neural correlates of System-1 and System-2 have also been studied (e.g., Tsujii & Watanabe, 2009). System-1 and System-2 are often used in fields such as psychology, philosophy, neuroscience, and artificial intelligence as a means for ontologizing the functional properties of human cognition.

Recently, however, this dual-system model has been criticized for lacking precision and conceptual clarity (Keren & Schul, 2009), leading to significant misconceptions (Pennycook et al., 2018; Houwer, 2019) and obscuring the dynamic complexities of psychological processes (Moors, 2016). One of the originators of dual-system theory stated that an important issue for future research is the problem that "current theories are framed in general terms and are yet to be developed in terms of their specific computational architecture" (Evans, 2003).

Following Dennett (1981) we argue that a computational description is essential for clarifying high level, psychological characterizations such as System-1 and System-2. At the time, Dennett received significant pushback on his view. However, we argue that it was too early in the development of computational models to fully appreciate the pragmatic value of his position.

In the spirit of this endeavour, Proust (2013) has argued that a more precise computational definition is needed to understand the role of System-1 and System-2 in metacognition. Proust defined these systems in terms of informational typologies (System-1 non-conceptual; System-2 conceptual). Similarly, Thomson et al. (2015) argued that the expert use of heuristics (System-1) could be defined in terms of instance based learning in ACT-R. In fact, there are numerous ways that cognitive models and cognitive architectures can and have been mapped onto the System-1 and 2 distinction. For example, dual-process approaches to learning have been instantiated within the CLARION architecture, modelling the interaction between implicit and explicit processes (Sun, Terry & Slusarz, 2005). System-1 and 2 have also been instantiated directly into the LIDA architecture (Faghihi et al., 2014).

While it is useful to work on modelling different aspects of System-1 and 2, the larger question is, in what sense is System-1 and 2 a valid construct? What are the necessary and sufficient conditions that precisely define System-1 and 2? And what are the cognitive and neural alignments to System-1 and System-2 (Evans, 2003)?

## The Common Model

The Common Model of Cognition, originally the 'Standard Model' (Laird et al., 2017) is a consensus architecture that integrates decades of research on how human cognition functions computationally. The Common Model represents a convergence across cognitive architectures regarding the modules and components necessary for biological and artificial intelligence. These modules are correlated with their associated brain regions and verified through neuroscience (Steine-Hanson et al., 2018). Neural evidence strongly supports the Common Model as a leading candidate for modeling the functional organization of the human brain (Stocco et al., 2021).

The computational processes of the Common Model are categorized into five components — working memory, perception, action, declarative memory, and procedural memory. Procedural memory is described as a production system which contains units called production rules (or 'productions'). The production system interacts with different modules through working memory represented as buffers. While these components are implemented differently among Common Model-type architectures, they describe a common functionality across implementations.

## System-1

Researchers generally describe System-1 by using a constellation of characteristics. Specifically, System-1 is described as fast, associative, emotional, automatic, and not requiring working memory (Kahneman, 2011; Evans, 2003; Strack & Deutsch, 2004). System-1 is considered to be evolutionary old and present within animals. It is composed of biologically programmed instinctive behaviours and operations that contain innate modules of the kind put forth by Fodor (1983). System-1 is not comprised of a single system but is an assembly of sub-systems that are largely autonomous (Stanovich & West, 2000). Automatic operations are usually described as involving minimal or no effort, and without a sense of voluntary control (Kahneman, 2011). Researchers generally agree that System-1 is made of parallel and autonomous subsystems that output only their final product into consciousness (often as affect), which then influences human decision-making (Evans, 2003). This is one reason the system has been called "intuitive" (Kahneman, 2003).

System-1 relies on automatic processes and shortcut strategies called heuristics — problem solving operations or rule of thumb strategies (Simon, 1955). The nature of System-1 is often portrayed as non symbolic, and has been associated with reinforcement learning (Barto et al., 1981) and neural networks (McLeod, 1998). Affect is integral to System-1 processes (Mitchell, 2011). Affect based heuristics result from an individual evaluating a stimulus based on their likes and dislikes. In more complex decision-making, it occurs when a choice is either weighed as a net positive (with more benefits than costs), or as net negative (less benefits than costs) (Slovic et al., 2004).

System-1 can produce what are called "cognitive illusions" that can be harmful if left unchecked. For example, the 'illusion of validity' is a cognitive bias in which individuals overestimate their ability to accurately predict a data set, particularly when it shows a consistent pattern (Kahneman & Tversky, 1973). Biases and errors of System-1 operate automatically and cannot be turned off at will. However, they can be offset by using System-2 to monitor System-1 and correct it.

## System-1 in the Common Model

System-1 can be associated with the production system which is the computational instantiation of procedural memory in the Common Model (Singley & Anderson, 1989). Procedural knowledge is represented as production rules ("productions") which are modeled after computer program instructions in the form of condition-action pairings. They specify a condition that, when met, will perform a prescribed action. A production can also be thought of as an *if-then* rule (Anderson, 1993). *If* it matches a condition, *then* it fires an action. Production rules transform information to resolve problems or complete a task, and are responsible for state-changes within the system. Production rules fire automatically off of conditions in working memory buffers. Their automaticity is due to the fact that they are triggered without secondary evaluation. Neurologically, production rules correlate with the 50ms decision timing in the basal ganglia (Stocco, Lebiere, & Anderson, 2010). The production system can enact reinforcement learning in the form of utility learning, where faster or more useful productions are rewarded and are more likely to be used later (Anderson, 1993). In a similar way, problem solving heuristics can be implemented as production rules (Payne et al., 1988).

The Common Model production system has many of the properties associated with System-1 such as being fast, automatic, implicit, able to implement heuristics, and reinforcement learning. However, the Common Model declarative memory system also has some of the properties associated with System-1. Specifically, associative learning and the ability to implement heuristics that leverage associative learning (Thomson et al., 2015). Here, it is important to understand that the Common Model declarative memory cannot operate without the appropriate productions firing, and without the use of buffers (working memory). Therefore, from a Common Model perspective, System-1 minimally involves productions firing based on buffer conditions, but can also involve productions directing declarative memory retrieval, which also relies on buffers. Based on this, System-1 cannot be defined as being uniquely aligned with either declarative or procedural memory. System-1 activity must involve production rules and buffers, and can also involve declarative knowledge.

## System-2

Researchers generally view System-2 as a collection of cognitive properties, characterized as slow,

propositional, rational, effortful, and requiring working memory (Kahneman, 2011; Strack & Deutsch, 2004; Frankish 2010). System-2 involves explicit propositional knowledge that is used to guide decision-making (Epstein & Pacini, 1999). Propositional knowledge is associated with relational knowledge (Halford, Wilson, & Phillips, 2010) which represents entities (e.g.: John and Mary), the relation between them (e.g.: loves) and the role of those entities in that relation (e.g.: John loves Mary). Higher level rationality in System-2 is also said to be epistemically committed to logical standards (Tsujii & Watanabe, 2009). System-2 processes are associated with the subjective experiences of agency, choice, and effortful concentration (Frankish, 2010)**.** The term "effortful" encompasses the intentional, conscious, and more strenuous use of knowledge in complex thinking. Higher level rationality is considered responsible for human-like reasoning, allowing for hypothetical thinking, long-range planning, and is correlated with overall measures of general intelligence (Evans, 2003).

Researchers have studied various ways in which System-2's effortful processes can intervene in System-1 automatic operations (Kahneman, 2003). Ordinarily, an individual does not need to invoke System-2 unless they notice that System-1 automaticity is insufficient or risky. System-2 can intervene when the anticipated System-1 output would infringe on explicit rules or potentially cause harm. For example, a scientist early in their experiment may notice that they are experiencing a feeling of certainty. System-2 can instruct them to resist jumping to conclusions and to gather more data. In this sense, System-2 can monitor System-1 and override it by applying conceptual rules.

## System-2 in the Common Model

Laird (2020) draws on Newell (1990), Legg and Hutter (2007) and others to equate rationality with intelligence, where "an agent uses its available knowledge to select the best action(s) to achieve its goal(s)." Newell's Rationality Principle involves the assumption that problem-solving occurs in a problem space, where knowledge is used to navigate toward a desired end. As Newell puts it, "an agent will use the knowledge it has of its environment to achieve its goals" (1982, p. 17). The prioritizing of knowledge in decision-making corresponds with the principles of classical computation involving symbol transformation and manipulation.

The Common Model architecture fundamentally distinguishes between declarative memory and procedural memory. This maps roughly onto the distinction between explicit and implicit knowledge — where declarative knowledge can be made explicitly accessible in working memory, procedural knowledge operates outside of working memory and is inaccessible. However, declarative knowledge can also function in an implicit way. The presence of something within working memory does not necessarily mean it will be consciously accessed (Wallach & Lebiere, 2003).

Higher level reasoning involves the retrieval of 'chunks', representing propositional information, into buffers (working memory) to assist in calculations and problem-solving operations. This appears to correlate with what System-2 researchers describe as "effortful", as this requires more computational resources (i.e., more productions) to manage the flow of information through limited space in working memory (buffers). As Kahneman points out, System-1 can involve knowledge of simple processes such as 2+2=4. However, more complex operations such as 17x16 require calculations that are effortful, a characteristic that is considered distinctive of System-2 (Kahneman, 2011).

Effort, within the Common Model, involves greater computational resources being allocated toward a task. Moreover, the retrieval and processing of declarative knowledge requires more steps and more processing time when compared to the firing of productions alone. This longer retrieval and processing time can also account for the characteristic of "slow" associated with System-2.

## Emotion in System-1 and 2

Emotion and affect plays a vital role in the distinction between System-1 and System-2 processes (Chaiken & Trope, 1999; Kahneman, 2011). Decisions in System-1 are largely motivated by an individual's implicit association of a stimulus with an emotion or affect (feelings that something is bad or good). Behavior motivated by emotion or affect is faster, more automatic, and less cognitively expensive. One evolutionary advantage of these processes is that they allow for split-second reactions that can be crucial for avoiding predators, catching food, and interacting with complex and uncertain environments.

Emotions can bias or overwhelm purely rational decision processes, but they can also be overridden by System-2 formal rules. While emotions and affect have historically been cast as the antithesis of reason, their importance in decision-making is being increasingly investigated by researchers who give affect a primary role in motivating decisions (e.g., Zajonc, 1980; Barrett & Salovey, 2002). Some maintain that rationality itself is not possible without emotion, as any instrumentally rational system must necessarily pursues desires (Evans, 2012).

## Emotion in the Common Model

Feelings and emotions have strong effects on human performance and decision-making**.** However, there is considerable disagreement over what feelings and emotions are and how they can be incorporated into cognitive models. However, while philosophical explanations of affect have been debated, functional accounts of emotions and feelings within cognitive models have been built. Emotions have been modeled as amygdala states (West & Young, 2017), and somatic markers as emotional tags attached to units of information (Domasio, 1994). In Sigma models, low-level appraisals have been modeled as architectural self-reflections on factors such as expectedness, familiarity, and desirability (Rosenbloom, et al., 2015). Core affect theory has been modeled in ACT-R to demonstrate how an agent may prioritize information using emotional valuation **(**Juvina, Larue & Hough, 2018). Also, feelings have also been modelled by treating them as non propositional representations in buffers or "metadata" (West & Conway-Smith, 2019).

Overall, the question of how to model emotion in the Common Model remains unresolved. However, as indicated in the research above, emotion has multiple routes for interacting with cognition in the Common Model.

## Effort in System-1 and 2

The concept of "effort" makes up a significant and confusing dimension of System-1 and System-2. While it is mainly associated with System-2 rationality, a precise definition of "effort" remains elusive and is largely implicit in discussions of System-1 and 2. Because System-2 is considered to have a low processing capacity, its operations are associated with greater effort and a de-prioritizing of irrelevant stimuli (Stanovich, 1999).

Effort can be associated with complex calculations in System-2 to the extent that it taxes working memory. Alternatively, effort can be associated with System-2's capacity to overrule or suppress automatic processes in System-1 (Kahneman, 2011). For example, various System-1 biases (such as the "belief bias") can be subdued by instructing people to make a significant effort to reason deductively (Evans, 1983). The application of formal rules to "control" cognitive processes is also called metacognition — the monitoring and control of cognition (Flavell, 1979; Fletcher & Carruthers, 2012). Researchers have interpreted metacognition through a System-1 and System-2 framework (Arango-Muñoz, 2011; Shea et al., 2014). System-1 metacognition is thought to be implicit, automatic, affect-driven, and not requiring working memory. System-2 metacognition is considered explicit, rule-based, and relying on working memory.

While the concept of "effort" is considered to be the monopoly of System-2, a computational approach suggests that effort is a continuum — with low effort cognitive phenomena being associated with System-1, and high effort cognitive phenomena being associated with System-2.

## Effort in the Common Model

The Common Model helps to elucidate how "effort" can be present in System-1 type operations in the absence of other System-2 characteristics. While neither dual-system theories nor the Common Model contain a clear definition of "effort", computational characteristics associated with effort can be necessary to System-1. For instance, "effort" is often associated with the intense use of working memory. However, the Common Model requires working memory (along with its processing limitations) for both System-1 and System-2 type operations. There is no reason why System-1 should necessarily use less working memory than System-2 in the Common Model. Instead, it would depend on the task duration and intensity.

System-1 and System-2 metacognition can also be clarified by importing Proust's (2013) more precise account. Proust attempted to elucidate these two systems by claiming that they should be distinguished by their distinctive informational formats (System-1 non-conceptual; System-2 conceptual). In this sense, System-1 metacognition can exert effortful control while simultaneously being implicit and non conceptual. For example, consider a graduate student attending a conference while struggling not to fall asleep. An example of System-1 metacognition would involve the context implicitly prompting them to feel nervous, noticing their own fatigue, and then attempting to stay awake. This effort is context-driven, implicit, non conceptual, and effortful. Alternatively, System-2 metacognition can exert effort by way of explicit concepts, as in the case of a tired conference-attendee repeating the verbal instruction "try to focus". Either of these scenarios could be modelled using the Common Model, and to reiterate, there is little reason why System-1 should require less effort.

Another way to think about effort is in terms of the expense of neural energy. In this sense, effort can be viewed as the result of greater caloric expenditure in neurons. The neural and computational dynamics responsible for the effortful control of internal states have shown to be sensitive to performance incentives (Egger et al., 2019). Research also indicates that the allocation of effort as cognitive control is dependent on whether a goal's reward outweighs its costs (Shenhav, et al., 2017). Both of these relate to reinforcement learning, which is associated with System-1.

Examining this question through the Common Model suggests that "effort" is not traditionally well defined, nor is it the sole privy of System-2. Rather, effort can be involved in processes characteristic of both System-1 and System-2.

## Conclusion

The Common Model sheds light on the specific mechanisms that give rise to the general traits associated with System-1 and System-2. Interpreting System-1 and System-2 within the Common Model results in our concluding that the "alignment assumption" (that the two systems are opposites) is a false dichotomy. There are, of course, cases where all properties of System-1 and System-2 are cleanly bifurcated on either side. However, between these two extremities lies a spectrum where the characteristics are mixed. Few, if any, of these properties are 'necessary and sufficient' to be sharply distinctive of either.

Evidence for this is as follows:

1. *System-2 is grounded in System-1.* While System-1 depends on procedural memory, so too does System-2. System-2 cannot operate separately due to the architectural constraints of the Common Model. Even if a System-2 process were primarily driven by declarative knowledge, it would still require System-1 procedural knowledge to be retrieved and acted upon.

2. *System-1 and System-2 characteristics are often mixed as they routinely act together.* System-2 goal-directed rationality often requires affect in the from of a desired end. Also, System-2 rationality is subject to System-1 affective biases.

3. *Both System-1 and System-2 require working memory.* While conventional views claim that System-1 does not require working memory, the constraints of the Common Model necessitate it. Production rules (procedural knowledge) are activated by the content of buffers (working memory) and hence are required by both systems.

4. *Effort can be directed toward both System-2 rationality and System-1 metacognitive control.* The effortful allocation of cognitive resources in System-1 can be based on an implicit cost-benefit analysis.

Regardless of whether one adopts the Common Model architecture, researchers should be cautious of assuming the System-1 and System-2 dichotomy within their work. The framework is far from settled and deep issues continue to be unresolved. Questions remain as to whether System-1 and System-2 constitute an ontology or a convenient epistemology.

Since before Descartes, substance dualism has continually been reimagined as mind and soul, reason and emotions, and opposing modes of thought. These have been expressions of the human species' attempt to make sense of our own minds, its processes, and how this understanding maps onto our personal experience. Clearly, System-1 and System-2 captures something deeply intuitive about the phenomenology of cognition. However, as we have discussed Kahneman's System-1 biases it may be worth asking — is System-2 a System-1 illusion? That is, do we assume the existence of System-2 simply because we so often act as if it exists?

By situating System-1 and System-2 within the Common Model of Cognition, we have attempted to bring light to this subject by clarifying its underlying mechanisms, misconceptions, and the base components needed for future research.